\documentclass[10pt,twocolumn,letterpaper]{article}

\usepackage{iccv}
\usepackage{times}
\usepackage{epsfig}
\usepackage{graphicx}
\usepackage{amsmath}
\usepackage{amssymb}
\usepackage{algorithm2e}
\usepackage{subfigure}
\usepackage{algpseudocode}

\usepackage[breaklinks=true,bookmarks=false]{hyperref}

\iccvfinalcopy 

\def\iccvPaperID{9809} 

\ificcvfinal\fi

\begin{document}

\title{Bayesian Eye Tracking\vspace{-0.2in}}


\author{\small Qiang Ji and Kang Wang\\
\small Rensselaer Polytechnic Institute\\
{\tt\small jiq@rpi.edu}
}

\maketitle
\ificcvfinal\thispagestyle{empty}\fi

\maketitle

\begin{abstract}
Model-based eye tracking has been a dominant approach for eye gaze tracking because of its ability to generalize to different subjects, without the need of any training data and eye gaze annotations. Model-based eye tracking, however, is susceptible to eye feature detection errors, in particular for eye tracking in the wild. To address this issue, we propose a Bayesian framework for model-based eye tracking.  The proposed system consists of a cascade-Bayesian Convolutional Neural Network (c-BCNN) to capture the probabilistic relationships between eye appearance and its landmarks, and a geometric eye model to estimate eye gaze from the eye landmarks.  Given a testing eye image, the Bayesian framework can generate, through Bayesian inference, the eye gaze distribution without explicit landmark detection and model training, based on which it not only estimates the most likely eye gaze but also its uncertainty. Furthermore, with Bayesian inference instead of point-based inference, our model can not only generalize better to different sub-jects, head poses, and environments but also is robust to image noise and landmark detection errors.  Finally, with the estimated gaze uncertainty, we can construct a cascade architecture that allows us to progressively improve gaze estimation accuracy.  Compared to state-of-the-art model-based and learning-based methods, the proposed Bayesian framework demonstrates significant improvement in generalization capability across several benchmark datasets and in accuracy and robustness under challenging real-world conditions.
\end{abstract}


\section{Introduction}
Eye tracking is to identify people's focus of attention(or line of sight) in the 3D space or on the 2D screen. Applications of eye tracking range from marketing, psychology, user behavior research to gaming, human-computer interaction, medical diagnosis, etc. 

Recent gaze estimation methods can be divided into two categories: 1) learning-based and 2) model-based.  Learning-based methods \cite{Tan2002,lu2011inferring,Valenti2012,Yiu-Ming2015,zhang15_cvpr,Krafka2016} aim at learning a mapping function from the input image to eye gaze. The most popular mapping function nowadays is the convolutional neural network and its variants.  Despite their impressive performance on benchmark datasets, the major issue with learning-based methods is that they cannot generalize well  and that they require significant amount training data with eye gaze annotations to perform well.  Accurate eye gaze annotation is a tedious, expensive, and time-consuming process and it often requires specialized equipment
\cite{wood2017gaze} .  
Model-based methods leverage on a geometric model representing the anatomical structure of eye to estimate eye gaze. The idea is to first detect 2D eye or facial landmarks, then reconstruct the 3D location of eyeball center and pupil center from 2D observations. Finally, 3D eye gaze can be estimated by connecting the eyeball center and pupil center.  Early efforts were focused on designing new geometric models \cite{beymer,Heinzmann1998,Ishikawa2004, Eizenman_general, kinect_EyeModel, Shih,Chen2008, kinect_xiong, Hirotake2008,Vicente2015,wang2017real} to better recover 3D information from 2D observations. Recently, researchers are more interested in improving the accuracy of eye/facial landmark detection \cite{kinect_Mora3,park2018deep,wang2018hierarchical,park2018learning} with the powerful deep learning tools. While training-free and generalizabale, the model-based approach often cannot produce accurate eye gaze estimation due to inaccuracies with eye landmark detection, in particular for real world eye tracking, unless special hardware (e.g IR light) is used.
\begin{figure*}[t]
  \centering  \includegraphics[width=1.0\linewidth]{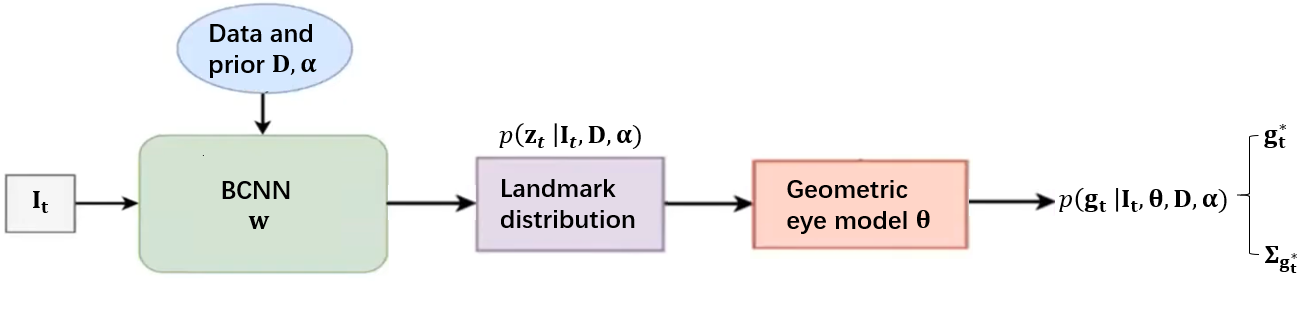}
  \caption{Overview of the proposed Bayesian eye tracking framework. The Bayesian Convolutional Neural Network (BCNN) captures the posterior distribution of model parameters from data $\mathbf{D}$ and prior $\boldsymbol \alpha$. Eye landmark distribution $p(\mathbf{z}_t |\mathbf{I}_t, \mathbf{D}, \boldsymbol \alpha)$ of a testing image $\mathbf{I}_t$ can be estimated with BCNN, which is then used to estimate the gaze distribution $p(\mathbf{g}_t |\mathbf{I}_t, \mathbf{D}, \theta, \boldsymbol \alpha)$ given a geometric eye model with parameter $\boldsymbol \theta$. And we can further compute optimal gaze prediction $\mathbf{g}^*_t$ and uncertainty $\mathbf{\Sigma_{g_t^*}}$ from gaze distribution.
  } 
  \label{fig:BET_Overview}
\end{figure*}

To overcome these limitations, we introduce a Bayesian framework for model-based eye tracking without explicit eye landmark detection. 
 As shown in Fig.~\ref{fig:BET_Overview}, the proposed method consists of a Bayesian Convolutional Neural Network (BCNN) module
  which captures the relationship between eye appearance and eye landmarks, and a geometric model which relates eye landmarks to eye gaze. The two modules are jointly formulated through a Bayesian probabilistic framework, and eye gaze can be effectively estimated through a Bayesian inference. With the Bayesian inference, eye landmark predictions on an input eye image can be performed by an ensemble of BCNN models drawn from their posterior distributions.  The landmarks generated by each BCNN model are then used to estimate eye gaze for the input image, yielding a distribution of eye gaze. By modeling the landmark distribution and gaze distribution,  the proposed model  can generalize  to different subjects, head poses or new environments and yield robust performance even under  noisy images.  Furthermore,  our model also associates a level of uncertainty with its gaze prediction. Uncertainty is essential to understand what deep networks cannot do and
avoid over-confident predictions. Finally, by feeding the uncertainty information from current stage to next stage with the cascade architecture, we can further improve the model performance. To summarize, we make following novel contributions:

\begin{itemize}
\item We introduce a unified Bayesian framework that combines a model-based eye tracking with a Bayesian convolutional neural network to allow robust and generalizable  model-based eye tracking .
\item With  Bayesian inference, our model avoids explicit landmark detection and requires no eye gaze annotations and no model training.
\item The model produces not only eye gaze predictions but also their uncertainties, which can be incorporated into  a cascade framework to progressively improve gaze estimation performance.
\end{itemize}

\section{Related work}
In this section, we focus on reviewing recent work on model-based eye gaze tracking, in particular those aimed at
improving eye feature detection and eye tracking generalization performance. In \cite{wang2017real}, the authors proposed to estimate the 3D eye gaze given 2D facial landmarks. They introduce a 3D eye-face deformable model that relates the 3D eyeball center with 3D rigid facial landmarks. Given the detected 2D facial landmarks and the eye-face model, they can effectively recover the 3D eyeball center and obtain the final gaze direction. In \cite{park2018learning}, the authors proposed to recover the gaze direction with eye region landmarks. They first used a deep network (Hourglass) to detect eye region landmarks. Next they leveraged on the proposed geometric eye model to estimate the eye gaze direction given the detected eye landmarks. Both methods require accurate facial and eye region landmark detections to perform well. 
In \cite{santini2018pure}, authors focused on robust eye feature detection in pervasive scenarios. They introduced the Pupil Reconstructor(PuRe), which is mainly based on an edge segment selection. Their method also includes the confidence measure for the detected pupil. But PuRe fails to work on low-resolution eye images.
Sun, \emph{et al.} \cite{sun2015real} proposed a gaze estimation system that consists of three modules: gaze feature extraction, system calibration, and gaze estimation. The gaze features are fed into the gaze estimation module after the calibration procedure. During the gaze feature extraction, a parameterized iris model is used to locate iris center, which is robust against low-qulity eye images and large head movement. Wood, \emph{et al.} \cite{Wood2016} constructed a 3D deformabel dense eye model from the head scans. Their model includes shape and texture variations of facial eye region (eyebrow, eyelid, part of the nose) and eyeball (iris, sclera). By fitting the whole eye model instead of fitting landmarks, they can get the 3D position of eyeball center and pupil center simultaneously. But during fitting, they still used landmarks as regularization and set high weights to it. And their model is not robust when facing real world scenario, especially when occlusion occurs.
Wang \emph{et al.} \cite{wang2019generalizing}, proposed to leverage Bayesian inference during the adversarial learning to increase generalization for appearance-based gaze estimation. However, as a learning based instead of model-based, their method required a large amount of labeled training data under different variations and a complex training procedure; it hence cannot generalize well beyond the training data.

\section{Problem statement}

Given data $\mathbf{D} = \{\mathbf{I}_i, \mathbf{z}_i\}_{i=1}^N$, where $\mathbf{I}_i$ and $\mathbf{z}_i$ represent the image and corresponding landmarks, model prior $\boldsymbol \alpha$ and geometric model parameter $\boldsymbol \theta$, our goal is to estimate the gaze $\mathbf{g}_t$ for a testing image $\mathbf{I}_t$. 
We want to emphasize here that $\mathbf{D}$ contains no eye gaze annotations and that
its landmarks are not manually annotated but are generated by a third party landmark detector. 

We first introduce the deterministic model formulation, then we discuss our extension to the probabilistic formulation.
\subsection{Deterministic formulation}

For deterministic formulation, we first map eye image to its landmarks through a landmark detector, then estimate eye gaze with the geometric eye model from the landmarks: 
\begin{align}
\label{eq:non_bayesian_I2z}
 \mathbf{z}_t = f(\mathbf{I}_t; \mathbf{w}) \hspace{1in}
 \mathbf{g}_t = h(\mathbf{z}_t; \boldsymbol \theta)
\end{align}
The image to landmark mapping function $f(\cdot)$ is  obtained by a training process. 
$h(\cdot)$ encodes the eye model that maps eye/facial landmarks to the eye gaze analytically. The eye model parameter $\boldsymbol \theta$ include both the eye model and
personal eye parameters (e.g. eyeball radius) and they can be obtained through an optimization and
an offline personal calibration. 

\begin{figure}[h]
  \centering  \includegraphics[width=1\linewidth]{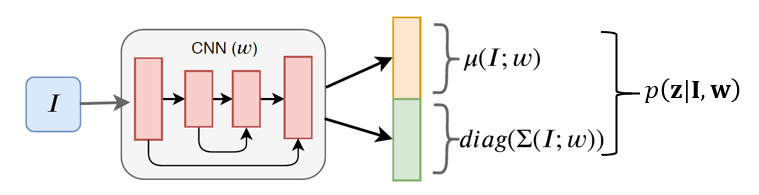}
  \caption{Illustration of BCNN.} 
  \label{fig:landmark_posterior}
\end{figure}
\subsection{Probabilistic formulation}

The deterministic formulation has several limitations. First, gaze estimation only relies on a single set of parameter $\mathbf{w}$, which may not work well when data comes from different subjects, head poses or environments or noisy eye images. Second, it requires to explicitly detect the landmark points for model-based gaze estimation.  Gaze estimation hence is sensitive to landmark detection errors, and it does not have a way to characterize the errors of the landmark detection.

To address these limitations, we propose to employ a Bayesian convolutional neural network (BCNN) instead of a conventional 
convolutional neural network (CNNs) to capture the probabilistic relationships between an input eye image and the eye/facial landmarks. 
With Bayesian inference, BCNN can produce an ensemble of CNNs by sampling its parameter posterior 
$p(\mathbf{w}|\mathbf{D}, \alpha)$. The ensemble of CNNs produce a distribution of landmarks, which, after fed into the geometric eye model, produces a
distribution of eye gaze for the input eye image. By using landmark distribution instead of explicit landmark detection with one model, we can improve both eye gaze estimation accuracy, robustness, and its generalization.  
Specifically, the proposed Bayesian eye tracking can be formulated as follows:
\begin{small}
\begin{align}
\mathbf{g}_t^* &= \arg \max_{\mathbf{g}_t} p(\mathbf{g}_t |\mathbf{I}_t, \mathbf{D}, \theta, \boldsymbol \alpha) \notag \\
\label{eq:int_z}
&= \arg \max_{\mathbf{g}_t} \int_{\mathbf{z}_t} p(\mathbf{g}_t, \mathbf{z}_t |\mathbf{I}_t, \mathbf{D}, \theta, \boldsymbol \alpha) d \mathbf{z}_t  \\
&= \arg \max_{\mathbf{g}_t} \int_{\mathbf{z}_t} p(\mathbf{g}_t | \mathbf{z}_t, \theta) p(\mathbf{z}_t |\mathbf{I}_t, \mathbf{D}, \boldsymbol \alpha) d \mathbf{z}_t \notag \\
\label{eq:int_w}
&= \arg \max_{\mathbf{g}_t} \int_{\mathbf{z}_t} p(\mathbf{g}_t | \mathbf{z}_t, \theta) \int_{\mathbf{w}} p(\mathbf{z}_t, \mathbf{w} |\mathbf{I}_t, \mathbf{D}, \boldsymbol \alpha) d \mathbf{w} d \mathbf{z}_t \\
\label{eq:gaze_overall}
&= \arg \max_{\mathbf{g}_t} \int_{\mathbf{z}_t} \underbrace{p(\mathbf{g}_t | \mathbf{z}_t, \theta)}_{\text{model}} \int_\mathbf{w} \underbrace{p(\mathbf{z}_t | \mathbf{I}_t, \mathbf{w})}_{\text{landmark dist}} \overbrace{p(\mathbf{w} | \mathbf{D}, \boldsymbol \alpha)}^{\text{parameter posterior}} d \mathbf{w}  d \mathbf{z}_t 
\end{align}
\end{small}


As can be seen in Eq.~\eqref{eq:int_z}, by integrating all possible landmarks $\mathbf{z}_t$, we avoid explicit landmark detection and it is more robust against noise. In addition, as in Eq.~\eqref{eq:int_w}, by integrating over all possible network parameters $\mathbf{w}$, we avoid point-based estimation, avoid over-fitting, and
are able to achieve better generalization across different subjects, head poses as well as different environments. 

For simplicity and better understanding, we first discuss the single-stage model (BCNN) in Sec.~\ref{sec-single-stage}. Then we discuss the extension to multi-stage (c-BCNN) in Sec.~\ref{sec-extension_cBCNN}.

\section{Single-stage gaze estimation}
Following the order in Eq.~\eqref{eq:gaze_overall}, we first discuss how to obtain 
the landmark distribution $p(\mathbf{z}_t | \mathbf{I}_t, \mathbf{w})$ for a testing image in Sec.~\ref{sec-uncertainty-based}. This is then followed by
discussing the parameter posterior $p(\mathbf{w} | \mathbf{D}, \boldsymbol \alpha)$ in Sec.~\ref{sec-posterior-inference} and then
the model-based gaze estimation $p(\mathbf{g}_t | \mathbf{z}_t, \theta)$ in Sec.~\ref{sec-model-based}. 
Finally, given these three terms, we summarize in Sec.~\ref{sec-overall-gaze} how we can approximately solve for Eq. \ref{eq:gaze_overall}.

\label{sec-single-stage}
\subsection{Landmark distribution and BCNN}
\label{sec-uncertainty-based}

We assume the landmark distribution term $p(\mathbf{z}_i|\mathbf{I}_i, \mathbf{w})$  or likelihood term follows a Gaussian distribution:
\begin{align}
\label{eq:feature_distribution}
p(\mathbf{z}_i|\mathbf{I}_i, \mathbf{w}) = \mathcal{N}(\mathbf{z}_i; \mu(\mathbf{I}_i; \mathbf{w}), \Sigma(\mathbf{I}_i; \mathbf{w}))
\end{align}
whose mean and covariance  matrix are outputted by the BCNN in Fig.~\ref{fig:landmark_posterior}. Note the inner part of the CNN is a U-Net(\cite{shrivastava2017learning}) like architecture. U-Net is proposed for segmentation tasks, and since the eye region can be segmented as skin, sclera and iris regions, landmarks therefore are easier to detect given the segmentation information.

In this work, the covariance matrix $\Sigma(\mathbf{I}_i; \mathbf{w})$ is assumed to be a diagonal matrix and hence the BCNN in Fig.~\ref{fig:landmark_posterior} only predicts the diagonal entries.

\subsection{Parameter posterior}
\label{sec-posterior-inference}

The parameter posterior $p(\mathbf{w}|\mathbf{D}, \boldsymbol \alpha)$ is proportional to the product of it’s likelihood and prior:
\begin{align}
\label{eq:w_posterior}
p(\mathbf{w}|\mathbf{D}, \boldsymbol \alpha) \approx p(\mathbf{D}| \mathbf{w}) p(\mathbf{w}|\boldsymbol \alpha) \notag \\= \prod_{i=1}^N p(\mathbf{z}_i|\mathbf{I}_i, \mathbf{w}) p(\mathbf{w}|\boldsymbol \alpha)
\end{align}
The likelihood part is generated by the BCNN as discussed in Section \ref{sec-uncertainty-based}.  For the prior term, we assume $p(\mathbf{w}|\boldsymbol \alpha) \sim \mathcal{N}(\mathbf{w}|\mathbf{0}, \sigma \mathbf{I})$, where $\sigma$ is a hyperparameter and is set to $1$ in this work. 


\subsection{Model-based gaze estimation from landmarks}
\label{sec-model-based}
\begin{figure}[t]
  \centering  \includegraphics[width=0.5\linewidth]{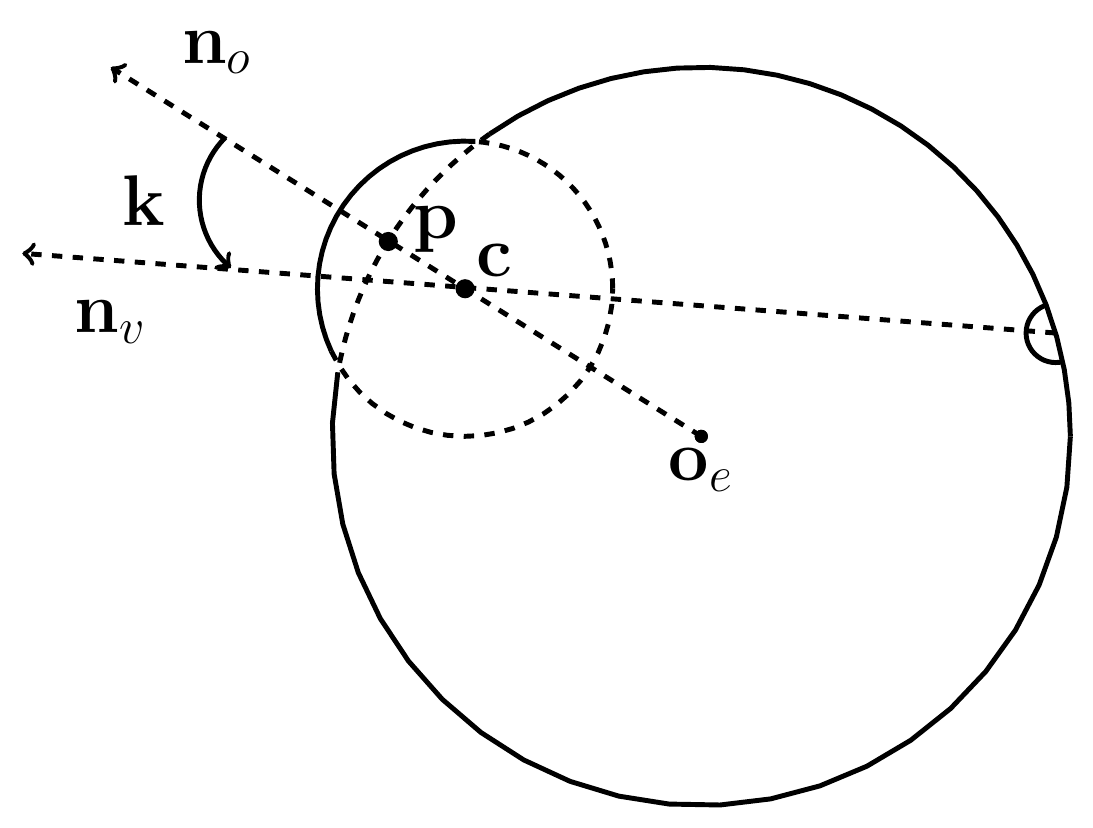}
  \caption{3D Geometric Eye Model.} 
  \label{fig:geometry_eye_model}
\end{figure}

The term $p(\mathbf{g}_t | \mathbf{z}_t, \theta)$ in Eq. \ref{eq:gaze_overall} encodes the model-based eye tracking. 
For model-based eye tracking, we focus on two recent models \cite{wang2017real,park2018learning} that only require a web camera. As two models employ the same principle, i.e., relating facial/eye landmarks to eye gaze through a geometric eye model, we focus on discussing \cite{wang2017real} below. They use an eye-face model, where 3D eye gaze can be estimated from facial landmarks (full face image is required), based on the classic two-sphere eye model as shown in Fig.~\ref{fig:geometry_eye_model}. 

According to the eye model, the true gaze is the visual axis $\mathbf{n}_v$ passing through cornea center $\mathbf{c}$ and fovea. An approximation of the true gaze is the optical axis which passes through eyeball center $\mathbf{o}_e$, cornea center $\mathbf{c}$ and pupil center $\mathbf{p}$. With the deformable 3D eye-face model, the authors first recover the 3D eyeball center $\mathbf{o}_e$ and the model parameters $\theta$ by minimizing the projection error of 3D eye face model to 2D landmarks. Next they compute the 3D pupil center $\mathbf{p}$ given observed 2D pupil center and the eyeball radius. Finally, they can recover the optical axis. Denote the 2D facial landmarks as $z_t$, gaze estimation can be summarized as $\mathbf{g}_t = h(\mathbf{z}_t; \boldsymbol\theta)$, where $\theta$ represents the model parameters. In addition, we also parameterize model-based gaze estimation in a probabilistic way, i.e.,
\begin{eqnarray}
p(\mathbf{g}_t|\mathbf{z}_t, \theta) = \mathcal{N}( h(\mathbf{z}_t;\theta), \Sigma_n) 
\end{eqnarray}
where $\Sigma_n$ is the noise with the model-based method and it can be estimated offline for each method.

\begin{figure}[t]
  \centering  \includegraphics[width=1.0\linewidth]{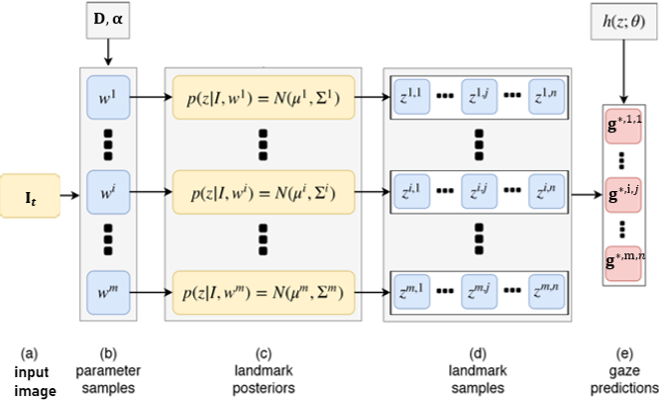}
  \caption{Overview of single-stage gaze estimation with Bayesian inference. The yellow box represents posterior distributions, while blue box represents samples drawn from the posteriors.} 
  \label{fig:Bayesian_inference}
\end{figure}

\subsection{Model-based gaze estimation via Bayesian inference}
\label{sec-overall-gaze}
Given the three terms in Eq.~\eqref{eq:gaze_overall}, we now introduce our method to solve the equation to obtain the best gaze and its uncertainty
for an input image.  
Solving eye gaze in Eq.~\eqref{eq:gaze_overall} directly is challenging because of the integration over both $\mathbf{z}$ and $\mathbf{w}$:
\begin{small}
\begin{align}
\mathbf{g}_t^* &= \arg \max_{\mathbf{g}_t} \int_{\mathbf{z}_t} p(\mathbf{g}_t | \mathbf{z}_t, \theta) \int_\mathbf{w} p(\mathbf{z}_t | \mathbf{I}_t, \mathbf{w}) p(\mathbf{w} | \mathbf{D}, \boldsymbol \alpha) d \mathbf{w} d \mathbf{z}_t  \notag 
\end{align}
\end{small}
We instead approximate the two integrals with sample averages. 

For the first integral over $\mathbf{w}$, because of the large number of parameters in $\mathbf{w}$,
we propose to 
employ the Stochastic Gradient Hamiltonian Monte Carlo (SGHMC) approach \cite{chen2014stochastic} to approximate the posterior. The key idea is to use Hamiltonian dynamics to produce a better sample proposal
and the mini-batch based
gradient update to scale up to large dataset.
Specifically, the Hamiltonians $H(\textbf{w}; \textbf{v})$ is a
function defined over the model parameter \textbf{w} and auxiliary
variable \textbf{v}:
\begin{align}
\label{eq-Hamiltonians}
H(\mathbf{w},\mathbf{v}) &\triangleq U(\mathbf{w}) + \frac{1}{2} \mathbf{v}^T M^{-1}\mathbf{v} \\ \nonumber
U(\mathbf{w})&\triangleq \sum_{i=i}^N\log p(\mathbf{z}_i|\mathbf{I}_i,\mathbf{w}) + \log p(\mathbf{w}|\boldsymbol \alpha)
\end{align}
where $M$ is set as a multiple of identity matrix for simplicity \cite{neal2011mcmc}. 
We compute the gradient update by simulating the Hamilton's equations. This can be implemented by a stochastic gradient update with momentum as follows:
\begin{align}
\label{eq-Hamiltonians-sgd}
\mathbf{w}_t &= \mathbf{w}_{t-1} + \mathbf{v}_{t}\\
\label{eq-Hamiltonians-sgd2}
\mathbf{v}_t &= (1-\beta) \mathbf{v}_{t-1} + \eta \nabla U(\mathbf{w}_{t-1}) + \mathcal{N}(0,2\eta \beta\mathbf{I})
\end{align}
where $\eta$ is the learning rate and $\beta$ is the friction term in SGHMC. Following Eq.~\eqref{eq-Hamiltonians-sgd}-\eqref{eq-Hamiltonians-sgd2}, we can map a state \textit{i.e.} value of ($\mathbf{w},\mathbf{v}$) at time $t$ to $t+1$. By repeating the updates we obtain samples of $\mathbf{w}$ at different $t$.
In experiment, we use a mini-batch of size $64$.
The samples of $\mathbf{w}$ are collected every $100$ updates after burn-in period of the Markov chain simulated by SGHMC.

Following the SGHMC method, we draw a sample $\mathbf{w}^i$ from $p(\mathbf{w} | \mathbf{D}, \boldsymbol \alpha)$ for a total of $m$ samples.
Eq. \ref{eq:gaze_overall} can hence be approximated by 
\begin{small}
\begin{align}
\label{eq:gaze_step1}
\mathbf{g}_t^* &\approx \arg \max_{\mathbf{g}_t} \frac{1}{m} \int_{\mathbf{z}_t} p(\mathbf{g}_t | \mathbf{z}_t, \theta) \sum_{i=1}^{m} p(\mathbf{z}_t | \mathbf{I}_t, \mathbf{w}^i)  d \mathbf{z}_t
\end{align}
\end{small}
Next we approximate the landmark integral by drawing samples $\mathbf{z}_t^{i, j}$ from $p(\mathbf{z}_t | \mathbf{I}_t, \mathbf{w}^i)$ (Eq.~\eqref{eq:feature_distribution}).
As we assume the covariance matrix for $p(\mathbf{z}_t | \mathbf{I}_t, \mathbf{w}^i)$ are diagonal, elements of  $\mathbf{z}_t$ are independent. Hence, we can sample each
landmark point independently from its mean and variance to generate $n$ samples corresponding each $\mathbf{w}^i$, 
\begin{align}
\label{eq:mode-of-mg}
\mathbf{g}_t^* &\approx \arg \max_{\mathbf{g}_t} \frac{1}{m*n}\sum_{j=1}^{n} \sum_{i=1}^{m}  p(\mathbf{g}_t | \mathbf{z}_t^{i, j}, \theta)   \\
\label{eq:approximate_gaze}
&\approx \frac{1}{m*n} \sum_{j=1}^{n} \sum_{i=1}^{m} \arg \max_{\mathbf{g}_t} p(\mathbf{g}_t | \mathbf{z}_t^{i, j}, \theta) 
\end{align}
Eq. \ref{eq:approximate_gaze} shows that the optimal gaze, given a set of landmarks, can be computed as
the average of the optimal gazes for each set of
landmarks.

Each set of landmark points $\mathbf{z}_t^{i, j}$ produces one corresponding gaze estimation $\mathbf{g}^{*,i,j}_t$, yielding a total of  $m\times n $ gaze estimation.
Given $m\times n $ gaze estimations, we can compute quantify the uncertainty of $\mathbf{g}_t^*$ using the sample covariance matrix as follows:
\begin{align}
\label{eq:uncertainty_gaze}
\Sigma_{\mathbf{g}_t^{*}} = \frac{1}{m\times n}\sum^n_{j=1}\sum^m_{i=1}{(\mathbf{g}_t^{*, i, j} - \mathbf{g}_t^{*})(\mathbf{g}_t^{*, i, j}- \mathbf{g}_t^{*})^T}
\end{align}

To better understand the overall algorithm, we summarize the process in Fig.~\ref{fig:Bayesian_inference} and Alg.~\ref{alg-gaze-estimation-flow}.

\begin{algorithm}[t]
\label{alg-gaze-estimation-flow}
 1. \textbf{Input:}  training data $\mathbf{D} = \{\mathbf{I}_i, \mathbf{z}_i\}_{i=1}^N$, model parameters $\theta$, prior $\boldsymbol \alpha$, testing image $\{\mathbf{I}_t\}_{t=1}^k$. \\
 2. \textbf{Output:} Estimated gaze $\{\mathbf{g}_t^*\}_{t=1}^k$. \\
 3. \textbf{Initialization:} network parameter $\mathbf{w}^0$, burn-in time $T$, collection interval $p$. \\
 4. Obtain $m$ samples $\{\mathbf{w}^i\}_{i=1}^{m}$ from the Markov chain:\\
 	\For{$iter \in \{1, ..., T + p*m\}$}
    {
    - Random sample a mini-batch of data from $\mathbf{D}$ \\
    - Perform one-step stochastic gradient update with the mini-batch data (Eq.~\eqref{eq-Hamiltonians-sgd} and Eq.~\eqref{eq-Hamiltonians-sgd2})\\
     $\mathbf{w}_t = \mathbf{w}_{t-1} + \mathbf{v}_{t}$, \\
     $\mathbf{v}_t = (1-\beta) \mathbf{v}_{t-1} + \eta \nabla U(\mathbf{w}_{t-1}) + \mathcal{N}(0,2\eta\beta \mathbf{I}) $\\ - After burn in time, collect $\mathbf{w}^i$ every $p$ updates.
    }
 5. Perform gaze estimation on testing images: \\
 	\For{$t \in \{1, ..., k\}$}
 	{
    - Sample $n$ set of landmarks for each $\mathbf{w}^{i}$: $\mathbf{z}_t^{i, j} \sim p(\mathbf{z}_t|\mathbf{I}_t, \mathbf{w}^{i}) \;\;\; \forall j \in \{1, ..., n\}$. \\
    - Model-based gaze estimation from landmarks:
    $\mathbf{g}_t^{*, i, j} = h(\mathbf{z}_t^{i, j}; \theta) \;\;\; \forall i, j$. \\
    - Final gaze estimation:
    $\mathbf{g}_t^{*} = \frac{1}{m*n} \sum_{i=1}^{m} \sum_{j=1}^{n} \mathbf{g}_t^{*, i, j}$\\
    $\Sigma_{\mathbf{g}_t^{*}} = \frac{1}{m\times n}\sum^n_{j=1}\sum^m_{i=1}{(\mathbf{g}_t^{*, i, j} - \mathbf{g}_t^{*})(\mathbf{g}_t^{*, i, j}- \mathbf{g}_t^{*})^T}$
    
    }
\caption{Overall gaze estimation flow}
\end{algorithm}
\label{sec-c-BCNN}
\begin{figure}[h]
  \centering  \includegraphics[width=0.8\linewidth]{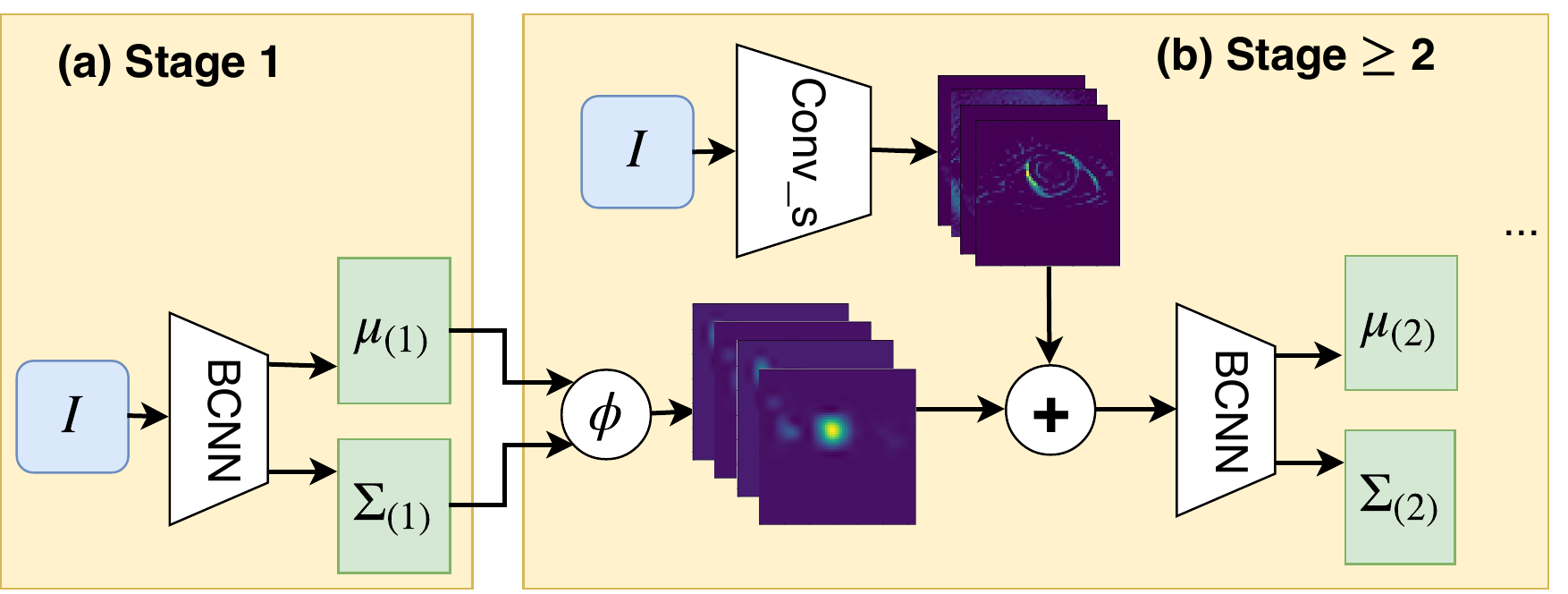}
  \caption{Cascade-Bayesian Convolutional Neural Network.} 
  \label{fig:BG_uncertainty}
\end{figure}
\section{Extension to multi-stage (c-BCNN)}
\label{sec-extension_cBCNN}
As the output of BCNN are the probability distribution of the landmark points, we can use them to improve landmark detection.  Moreover, several studies show that the cascade framework can yield better landmark predictions \cite{wei2016cpm}. We adopt the same idea of cascade framework with a modification that we incorporate the uncertainty information (Fig.~\ref{fig:BG_uncertainty}). Compared to \cite{wei2016cpm}, our method explicitly models the landmark distribution as a Gaussian distribution in Eq.~\eqref{eq:feature_distribution}, and predicts the mean and variance of the distribution, while \cite{wei2016cpm} predicts a deterministic landmark heatmap. In addition, \cite{wei2016cpm} uses artificially generated heatmap with fixed variance as supervision. Their fixed variance needs to be tuned carefully while our method only requires landmark locations as groundtruth without additional assumptions.

As can be seen in Fig.~\ref{fig:Bayesian_inference} (c), the $m$ samples of $\mathbf{w}$ produce $m$ posterior distributions of landmarks for the same image $\mathbf{I}$, from which we can compute both the epistemic and aleatoric uncertainties:
\begin{small}
\begin{align}
\label{eq:total-uncertainty}
\boldsymbol \mu_{(1)} &= \frac{1}{m} \sum_{i=1}^m \boldsymbol \mu_1^i \\
\boldsymbol \Sigma_{(1)} &= \underbrace{\frac{1}{m} \sum_{i=1}^m (\boldsymbol \mu_1^i - \boldsymbol \mu_{(1)})(\boldsymbol \mu_1^i - \boldsymbol \mu_{(1)})^T}_{\text{epistemic uncertainty}} +  \underbrace{\frac{1}{m} \sum_{i=1}^m \boldsymbol \Sigma_1^i}_{\text{aleatoric uncertainty}}
\end{align}
\end{small}
where $\boldsymbol \mu_{(1)}$ represents the average landmark predictions from the $m$ posterior samples $\mathbf{w}^i$, while $\boldsymbol \Sigma_{(1)}$ represents the total uncertainty about the prediction. Total uncertainty consists of the epistemic uncertainty term and the aleatoric uncertainty term. Epistemic uncertainty captures the covariance of the mean of the $m$ predictions. Intuitively, if the $m$ predictions are close to each other, we are more confident in the prediction, otherwise if the $m$ predictions are different from each other, we have high uncertainty about the prediction. Aleatoric uncertainty is the mean of the predicted covariance. It encodes the uncertainty about our input $\mathbf{I}$. It is high if $\mathbf{I}$ is different from the training data distribution (\textit{e.g.}, noise, outlier). 

Since total uncertainty captures the quality of the landmark predictions, we would like to incorporate uncertainty information to improve the performance. We propose to encode the total uncertainty into a probability map (same size as the feature map extracted from convolutional layers). The probability map encodes the probability of the landmark being at a particular pixel location. 

\begin{figure}[t]
  \centering  \includegraphics[width=0.8\linewidth]{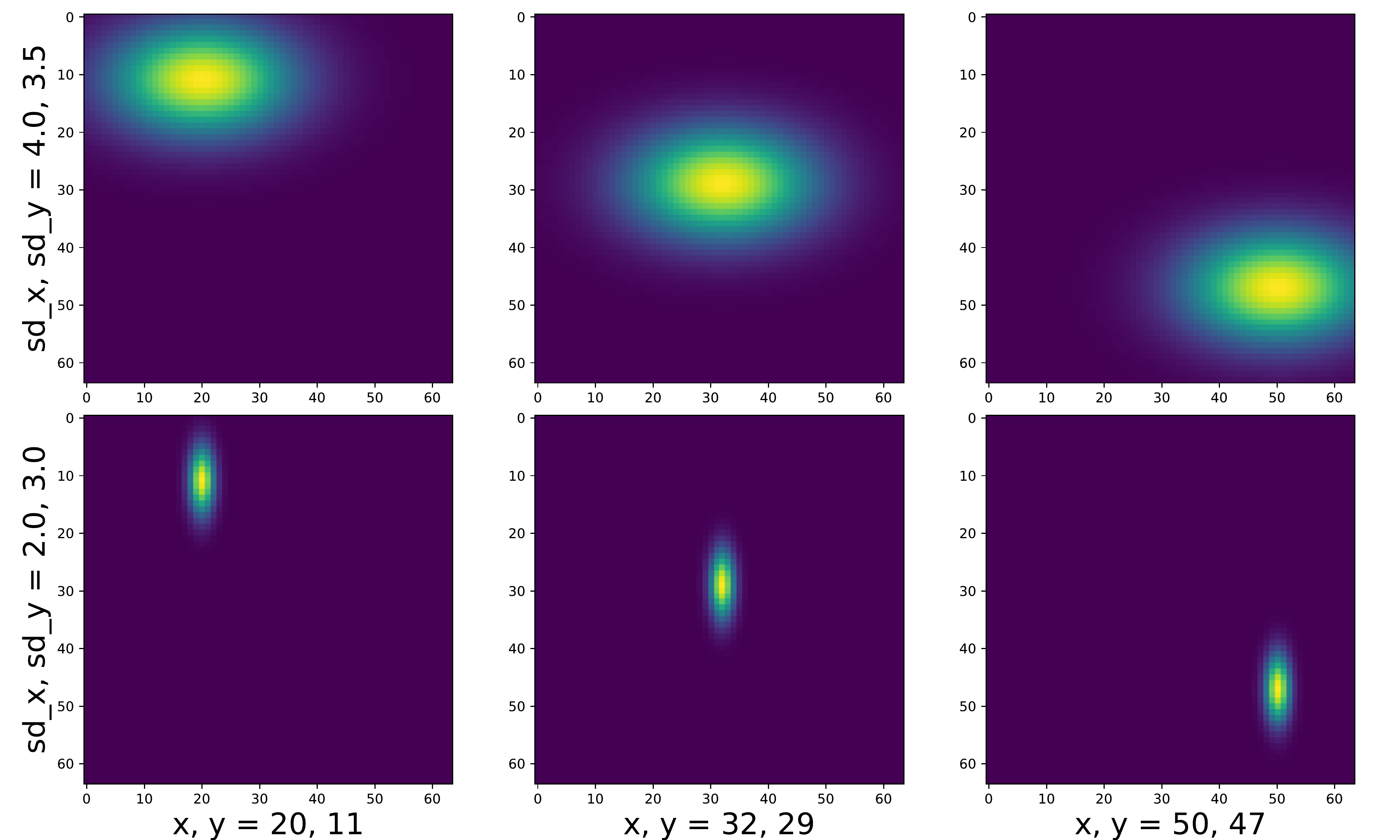}
  \caption{Example of probability maps given different predictions and total uncertainties. Feature map size is $64\times64$ pixels. Each row shows the same uncertainty at different landmark locations, each column shows the same landmark location with different uncertainties.} 
  \label{fig:vec2map}
  \vspace{-0.1in}
\end{figure}

To this end, we carefully design a probability map layer $
\boldsymbol\phi(\cdot)$, that converts $\boldsymbol \mu_{(1)}$ and $\boldsymbol \Sigma_{(1)}$ to a probability map $\mathbf{p} = \boldsymbol\phi(\boldsymbol \mu_{(1)}, \boldsymbol \Sigma_{(1)})$. The idea is to apply a Gaussian filter at location $\boldsymbol \mu_{(1)}$ with the total uncertainty $\boldsymbol \Sigma_{(1)}$ as covariance matrix. Let us consider the case for one landmark. Assume image is of size $h \times w$, $\boldsymbol \mu_{(1)} =(x, y)$, $diag(\boldsymbol \Sigma_{(1)}) = (\sigma_x, \sigma_y)$ (we assume independent dimensions of $\mathbf{z}$ and only use the diagonal entries) . We first create meshgrids $\mathbf{X}$ and $\mathbf{Y}$ with size $h \times w$, then the value of grid $(x,y)$ can be computed as follows:
\begin{align}
\label{eq:probability_map}
\mathbf{p} &= \boldsymbol\phi(\boldsymbol \mu_{(1)}, \boldsymbol \Sigma_{(1)}) = \boldsymbol\phi((x, y), (\sigma_x, \sigma_y)) \notag \\
&= \exp(-\frac{(\mathbf{X} - x)^2}{2\sigma_x^2}-\frac{(\mathbf{Y} - y)^2}{2\sigma_y^2})
\end{align}
To better understand how Eq.~\eqref{eq:probability_map} works, we provide several examples in Fig.~\ref{fig:vec2map}. Intuitively, if we have high uncertainty of the landmark prediction, the probability map will spread out (first row of Fig.~\ref{fig:vec2map}), and encourage the network to search around the neighborhood to refine the prediction. Similarly, network is encouraged to stay at current locations if we are confident on the prediction (peak regions as in second row of Fig.~\ref{fig:vec2map}). The complete c-BCNN is illustrated in Fig.~\ref{fig:BG_uncertainty}, the $Conv_s$ blocks are shared for Stage $\ge2$. The probability map and the feature map is concatenated and fed to the next BCNN block. With multiple stages, we can still use Alg.~\ref{alg-gaze-estimation-flow} to perform gaze estimation. The difference is that the $\boldsymbol \mu(\mathbf{I}; \mathbf{w})$ and $\boldsymbol \Sigma(\mathbf{I}; \mathbf{w})$ in the log-likelihood (Eq.~\eqref{eq:feature_distribution}) now is parameterized by the multi-stage CNN in Fig.~\ref{fig:BG_uncertainty}, where $\mathbf{w} = \{\mathbf{w}_{(1)}, ..., \mathbf{w}_{(k)}\}$ represents the parameters for all $k$ stages.

\section{Experiments and analysis}
We consider four benchmark datasets for our evaluation.
1) EyeDiap \cite{FunesMora_ETRA_2014}: The dataset contains different sessions (\textit{e.g.} HD camera VS VGA camera, discrete target VS floating target, small head motion VS large head motion). We use images captured by VGA camera with different head poses. 2) Columbia \cite{Smith2013}: The dataset has high quality images $5181\times3456$ from $55$ subjects. The $5880$ images are all used for evaluation. 3) UT \cite{Sugano2014}: The dataset contains $50$ subjects each with $8$ different head poses and $160$ gaze directions. All images are used for evaluations. 4) MPIIGaze \cite{zhang15_cvpr}: The dataset has $15$ subjects and there are head pose variations and lighting variations. We follow the same evaluation sittings to test on the evaluation subset from the dataset. 

The eye-region model is applied to the four datasets, while the eye-face model is only applied to EyeDiap and ColumbiaGaze since they provide with full face images and camera parameters (required by eye-face model). Gaze estimation error is measured by the angular error in degree.
Notice that these benchmark datasets do not provide landmark labels. For training data, we directly apply a third-party landmark detector \cite{gou2017coupled,gou2017joint} to generate landmarks (we use $11$ facial landmarks as in \cite{wang2017real} and $18$ eye landmarks as in \cite{park2018learning}). In spite of the inaccuracies with the landmark detector, we can also achieve comparable performance. For the cascade framework, we find that $6$ stages work best for facial landmarks and $3$ stages are sufficient for eye landmarks detection. All network inputs (face image/eye image) are resized to $184\times184$. For the convolutional blocks in each stage, we adopt a 4-layer U-Net like structure (Fig.~\ref{fig:BG_uncertainty}). The feature map size for the 4-layers are $92\times92$, $46\times46$, $46\times46$ and $92\times92$. For the geometric model in \cite{wang2017real}, we use the average 3D deformable eye-face model. We also use human-average eye related parameters (\textit{e.g.} eyeball radius is 12 mm, kappa angle is 5.0 and 1.2 degrees) for both eye-face model and eye-region model. For the Bayesian inference, we use $m=50$ parameter samples and $n=50$ landmark samples. 

\subsection{Evaluation of the Bayesian framework}
The proposed method consists of two major components: 1) Bayesian inference and 2) cascade framework. To evaluate contribution of Bayesian inference, we compare with two point estimators: Maximum Likelihood Estimation (MLE) and Maximum a Posterior (MAP) estimation. For MAP, we use the same prior $p(\mathbf{w}) = \mathcal{N}(\mathbf{0}, \mathbf{I})$ as our Bayesian inference. To evaluate the cascade framework, we consider onestage, two-stage and the final three-stage models. In addition, we perform cross subject (within-dataset) and cross-dataset experiments on four benchmark datasets to evaluate the two components.

\subsubsection{Contribution of Bayesian inference}
\begin{figure}[h]
  \centering  \includegraphics[width=0.9\linewidth]{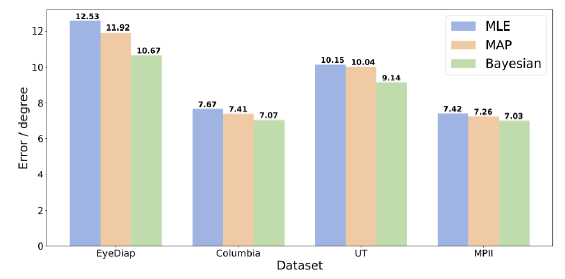}
  \caption{Within-dataset evaluations on four benchmark datasets} 
  \label{fig:within_dataset}
\end{figure}

\textbf{Within-dataset cross-subject performance:} The evaluation on four benchmark datasets are shown in Fig.~\ref{fig:within_dataset}. First of all, for the four benchmark datasets, the Bayesian inference outperforms the MLE and MAP point estimators. This demonstrates the importance of Bayesian inference in gaze estimation.

\vspace*{-\baselineskip}
Secondly, Bayesian inference is able to handle challenging datasets with low quality images. The Bayesian inference demonstrates more improvement in the two more challenging datasets (EyeDiap and UT) as MPIIGaze dataset has small head pose and subject variations while Columbia dataset has 4K resolution images and fixed head poses, This is because when the image distribution is complex, the corresponding landmark distribution is not accurate. This makes it hard for point estimators to cover the underlying distributions. On the other hand, explicitly modeling the landmark distribution and performing Bayesian inference allows us to leverage the distribution learned from training data and improves the estimation accuracy. Finally, average over the 4 datasets, Bayesian inference gives an improvement of $9.5\%$ over MLE and $6.8\%$ for MAP, demonstrating the effectiveness of the proposed Bayesian inference.

\begin{figure}[h]
  \centering  \includegraphics[width=0.9\linewidth]{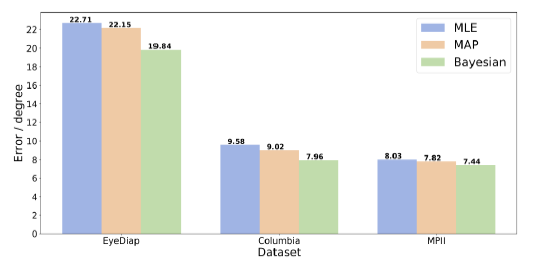}
  \caption{Cross-dataset evaluations on three benchmark datasets (best view in color). Training dataset is UT.} 
  \label{fig:cross_dataset}
\end{figure}

\textbf{Cross-dataset performance:}The cross-dataset experiments are shown in Fig.~\ref{fig:cross_dataset}, we obtain improvements with Bayesian inference for all three benchmark datasets. Cross-dataset evaluation suffers from the difference from training dataset distribution and testing dataset distribution. In general, point estimators only obtain one set of parametes, though it fits well to the training dataset, it cannot generate good results on testing datasets. With Bayesian inference and using multiple set of parameters, we are able to achieve improved results on cross-dataset evaluations. On average, Bayesian inference gives an improvement of $12.3\%$ over MLE and $9.0\%$ over MAP, which is also higher than the within-dataset experiments.

\textbf{Extreme-case evaluation:}To further demonstrate the advantage of Bayesian inference, We perform an extreme-case study where we add noises and occlusions on the testing images. In particular, we select UT as our training dataset, and MPII as our testing dataset. The experimental results are shown in Tab.~\ref{tab-extreme-case-study}, where the Gaussian noise standard deviation is applied to image with intensity from 0 to 255. The random occlusion is a squared black block, and the size of the square is the percentage of input image width. As we can see in the table, although MLE and MAP can still give reasonable estimation, Bayesian inference outperforms them in all scenarios, demonstrating its ability to handle images with poor qualities.
\begin{table}[h]
    \vspace{-0.1in}
    \renewcommand{\arraystretch}{1.2}
    \tabcolsep0.06in
  \begin{center}
    \caption{Extreme-case evaluations}
    \begin{tabular}{|l|c|c|c|c|c|c|} \hline 
    \ &\multicolumn{3}{|c|}{Gaussian Noise std} & \multicolumn{3}{|c|}{Random Occlusion size} \\ \hline 
     \ & 10 & 30 & 50 & $10\%$ & $20\%$ & $30\%$ \\ \hline 
     MLE\ & 9.21 & 10.77 & 12.57 & 8.93 & 11.25 & 14.36 \\ \hline 
     MAP\ & 8.48 & 10.15 & 11.69 & 9.12 & 10.79 & 14.23 \\ \hline 
     Bayesian\ & 7.97 & 9.23 & 10.26 & 8.24 & 9.57 & 13.07 \\ \hline 
    \end{tabular} 
    \label{tab-extreme-case-study} 
  \end{center}
  \vspace{-0.4in}
\end{table}

\subsubsection{Contribution of cascade framework}
We also perform the within-dataset and cross-dataset evaluations with different number of stages. The idea of cascade framework is that the prediction of the output will guide the searching in following stage. The major observations are: 1) more stages are always helpful and 2) the improvement from stage1 to stage2 is more significant than from stage2 to stage3. This is also the main reason we choose 3 stages for our c-BCNN. On average, the improvement of stage3 over stage1 for within-dataset is $14.7\%$, and for cross-dataset is $13.5\%$. This demonstrate the significance of the proposed cascade framework.
\vspace{-0.1in}
\subsubsection{Visualization of landmark detections}
\begin{figure}[t]
\subfigure{\centering  \includegraphics[width=1\linewidth]{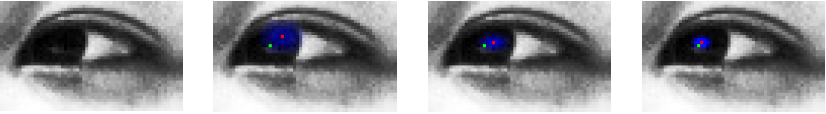}}
\subfigure{\centering  \includegraphics[width=1\linewidth]{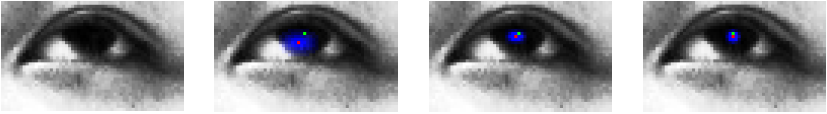}}
\subfigure{\centering  \includegraphics[width=1\linewidth]{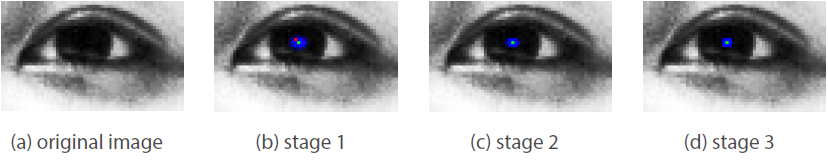}}
\caption{Examples of landmark detections across different stages. Green dots represent the groundtruth pupil center. Red and blue color represent the mean and variance of the predictions. Larger region of blue areas indicate high uncertainty about the prediction.}
\label{fig:cascade_stages}
\end{figure}

We also visualize the landmark detection results across different stages as in Fig.~\ref{fig:cascade_stages} (images from UT). In the first stage, the predicted mean (red dot) is far away from ground truth (green dot), the predicted variance (blue region) therefore is large, indicating high uncertainty. As we proceed with more stages, we can progressively improve the accuracy and reduce the uncertainty.
\subsection{Comparison with state of the art}
In this section, we summarize the experimental results of comparing our methods to SOTA methods.  We first compare with SOTA model-based methods,
and with SOTA learning-based methods.

\subsubsection{Model-based methods}
We compare with three recent model based methods \cite{wang2017real,park2018learning,wang2018hierarchical}
on three benchmark datasets. 
The within-dataset cross-subject experiments are shown in Tab.~\ref{tab-within-dataset}. For the eye-face model, we obtain a significant improvement ($16.2\%$) for the EyeDiap dataset, and slight better improvements for Columbia dataset. For the eye-region model, we achieve larger improvements on EyeDiap ($10.1\%$ compared with \cite{park2018learning},$29.6\%$ compared with \cite{wang2018hierarchical}), and similarly slight improvement on the two high quality datasets Columbia and MPII. Overall, this demonstrates that the proposed method can better generalize across subjects, especially when the distribution is complex and data is noisy. To compare the generalization performance, we also performance cross-dataset evaluation. Tab.~\ref{tab-cross-dataset} shows the results for cross-dataset evaluation. For fair comparison, we train the model with UT dataset and test on the remaining three datasets. We can see that we achieve a significant improvement on EyeDiap from 26.6 degrees to 20.6 degrees, showing the power for c-BCNN to generalize to a new dataset. For Columbia and MPII, we achieve a similar performance compared to competing approaches. The reason is that Columbia has high-quality images and the variation in MPII is relatively small. Overall, the results demonstrate better generalization capability of the proposed method.
\begin{table}[t]
  \begin{center}
    \caption{Within-dataset Cross-subject evaluation.}
    \begin{subtable}
    \centering
    \begin{tabular}{|l|c|c|c|c|} \hline 
      Eye-face model & EyeDiap & Columbia   \\ \hline 
      \cite{wang2017real} & 17.3 & 7.1  \\ \hline
      Proposed & \textbf{14.5} & \textbf{6.9}   \\ \hline
    \end{tabular}
    \end{subtable}
    \begin{subtable}
    \centering
    \begin{tabular}{|l|c|c|c|} \hline 
      Eye-region model & EyeDiap & Columbia & MPII   \\ \hline
      \cite{park2018learning} & 11.9 & 7.1 & - \\ \hline
      \cite{wang2018hierarchical} & 15.2 & -& 7.5 \\ \hline
      Proposed & \textbf{10.7} & \textbf{7.0} & \textbf{7.0} \\ \hline 
    \end{tabular}
    \end{subtable}
    \label{tab-within-dataset}  
  \end{center}
  \vspace{-0.2in}
\end{table} 
\begin{table}[t]
  \begin{center}
    \caption{Cross-dataset evaluation. Models are trained on UT dataset.}
    \begin{tabular}{|l|c|c|c|} \hline 
     Method & EyeDiap & Columbia & MPII  \\ \hline 
      \cite{wang2018hierarchical} & - & - & 7.7 \\ \hline
      \cite{park2018learning} & 26.6 & 8.3 & 8.7 \\ \hline
      Proposed & \textbf{20.6}  & \textbf{8.0} & \textbf{7.4}  \\ \hline
    \end{tabular} 
    \label{tab-cross-dataset} 
  \end{center}
  \vspace{-0.3in}
\end{table}

\subsubsection{Appearance-based methods}
Here, we only compare with SOTA learning-based methods on cross-datasets. We did not compare with SOTA methods for within-datasets 
because as a model-based method, our method does not require any training data with
groundtruth gaze annotations, while the learning-based methods all require a lot groundtruth gaze annotations, 
and hence the comparison is hence not fair. 

We follow a similar setting to train the model with UT dataset and test on MPIIGaze dataset. Current studies show that appearance-based methods (\cite{zhang15_cvpr,cheng2018appearance,zhang2017mpiigaze,fischer2018rt}) perform poorly on cross-dataset. We observe similar results as in Tab.~\ref{tab-cross-dataset-soa}. As model-based methods leverage on eye models, therefore the proposed method outperforms most appearance-based methods. The method in \cite{wang2019generalizing}, gain a similar cross-dataset performance. They apply Beyasian inference in the process of adversarial learning. However, their method require a large amount of labeled data under different variations and a complex training process. In contrast, our proposed model does not require any gaze label, model training, and even landmarks in the training data are generated through third-part landmark detectors.
\begin{table}[h]
\vspace{-0.1in}
  \begin{center}
    \caption{Cross-dataset evaluation with eye-region model.}
    \begin{tabular}{|l|c|c|c|c|c|c|} \hline 
     Method & \cite{zhang15_cvpr} & \cite{cheng2018appearance} & \cite{zhang2017mpiigaze} &\cite{fischer2018rt}& \cite{wang2019generalizing} & ours  \\ \hline
      Error & 13.9 & 8.8 & 9.8 & 8.9 & \textbf{7.4} & \textbf{7.4} \\ \hline
    \end{tabular} 
    \label{tab-cross-dataset-soa} 
  \end{center}
  \vspace{-0.2in}
\end{table}

\section{Conclusion}
We introduce a novel Bayesian framework to improve the generalization and robustness of the model-based eye tracking. It consists of a BCNN, which captures the probabilistic relationship between appearance and landmarks, and a geometric eye model to relate landmarks to gaze. The two components are unified by a Bayesian framework to avoid point estimation and the explicit landmark predictions. We further extend the single-stage model to multi-stage and incorporate uncertainty information to progressively improve the model performance. Experiments demonstrate better generalization and robustness with Bayesian inference and the cascade framework. Furthermore, our method outperforms SoA model-based eye tracking methods on challenging datasets, further demonstrating its generalization power.




{\small
\bibliographystyle{ieee}
\bibliography{ref}
}

\end{document}